\documentclass[times,final]{elsarticle}

\usepackage{arxiv}
\usepackage{graphicx}%
\usepackage{multirow}%
\usepackage{amsmath,amssymb,amsfonts}%
\usepackage{amsthm}%
\usepackage{mathrsfs}%
\usepackage[title]{appendix}%
\usepackage{xcolor}%
\usepackage{booktabs}%
\usepackage{subcaption}%
\usepackage{algorithm}%
\usepackage{algorithmicx}%
\usepackage{algpseudocode}%
\usepackage{listings}%
\usepackage{url}
\usepackage{float}
\usepackage{placeins}
\usepackage{multicol}
\usepackage[hidelinks]{hyperref}

\theoremstyle{thmstyleone}%
\theoremstyle{thmstyletwo}%

\theoremstyle{thmstylethree}%
\AtBeginDocument{%
  \def\thefnote{\myfnsymbol{fnote}}}
\makeatletter
\def\myfnsymbol#1{\expandafter\@myfnsymbol\csname c@#1\endcsname}
\def\@myfnsymbol#1{\ifcase #1\or $\dagger$\or $\#\#$\else \@ctrerr\fi}
\def\fntext[#1]#2{\g@addto@macro\@fnotes{%
   \refstepcounter{fnote}\elsLabel{#1}%
   \def\thefootnote{\thefnote}
   \global\setcounter{footnote}{\c@fnote}%
   \footnotetext{#2}}}
\makeatother

\begin{document}

\begin{frontmatter}

\title{The Endoscapes Dataset for Surgical Scene Segmentation, Object Detection, and Critical View of Safety Assessment: Official Splits and Benchmark}

\author[1]{Aditya \snm{Murali} \fnref{fn1}\corref{corresp}}
\author[1]{Deepak \snm{Alapatt}\fnref{fn1}}
\author[2,4]{Pietro \snm{Mascagni}\fnref{fn1}}
\author[2]{Armine \snm{Vardazaryan}}
\author[2]{Alain \snm{Garcia}}
\author[3]{Nariaki \snm{Okamoto}}
\author[4]{Guido \snm{Costamagna}}
\author[2]{Didier \snm{Mutter}}
\author[3]{Jacques \snm{Marescaux}}
\author[3]{Bernard \snm{Dallemagne}}
\author[1,2]{Nicolas \snm{Padoy}}

\cortext[corresp]{Corresponding author: \texttt{murali@unistra.fr}}
\fntext[fn1]{These authors contributed equally and share co-first authorship.}

\address[1]{ICube, University of Strasbourg, CNRS, Strasbourg, France}
\address[2]{IHU Strasbourg, Strasbourg, France}
\address[3]{Institute for Research against Digestive Cancer (IRCAD), Strasbourg, France}
\address[4]{Fondazione Policlinico Universitario A. Gemelli IRCCS, Rome, Italy}

\received{XXX}
\finalform{XXX}
\accepted{XXX}
\availableonline{XXX}
\communicated{XXX}

\begin{abstract}
This technical report provides a detailed overview of \textit{Endoscapes}, a dataset of laparoscopic cholecystectomy (LC) videos with highly intricate annotations targeted at automated assessment of the Critical View of Safety (CVS).
Endoscapes comprises 201 LC videos with frames annotated sparsely but regularly with segmentation masks, bounding boxes, and CVS assessment by three different clinical experts.
Altogether, there are 11090 frames annotated with CVS and 1933 frames annotated with tool and anatomy bounding boxes from the 201 videos, as well as an additional 422 frames from 50 of the 201 videos annotated with tool and anatomy segmentation masks.
In this report, we provide detailed dataset statistics (size, class distribution, dataset splits, etc.) and a comprehensive performance benchmark for instance segmentation, object detection, and CVS prediction.
The dataset and model checkpoints are publically available at \href{https://github.com/CAMMA-public/Endoscapes}{https://github.com/CAMMA-public/Endoscapes}.

\end{abstract}
\end{frontmatter}
\thispagestyle{empty}

\begin{figure*}[!h]
\centering
\includegraphics[width=\textwidth]{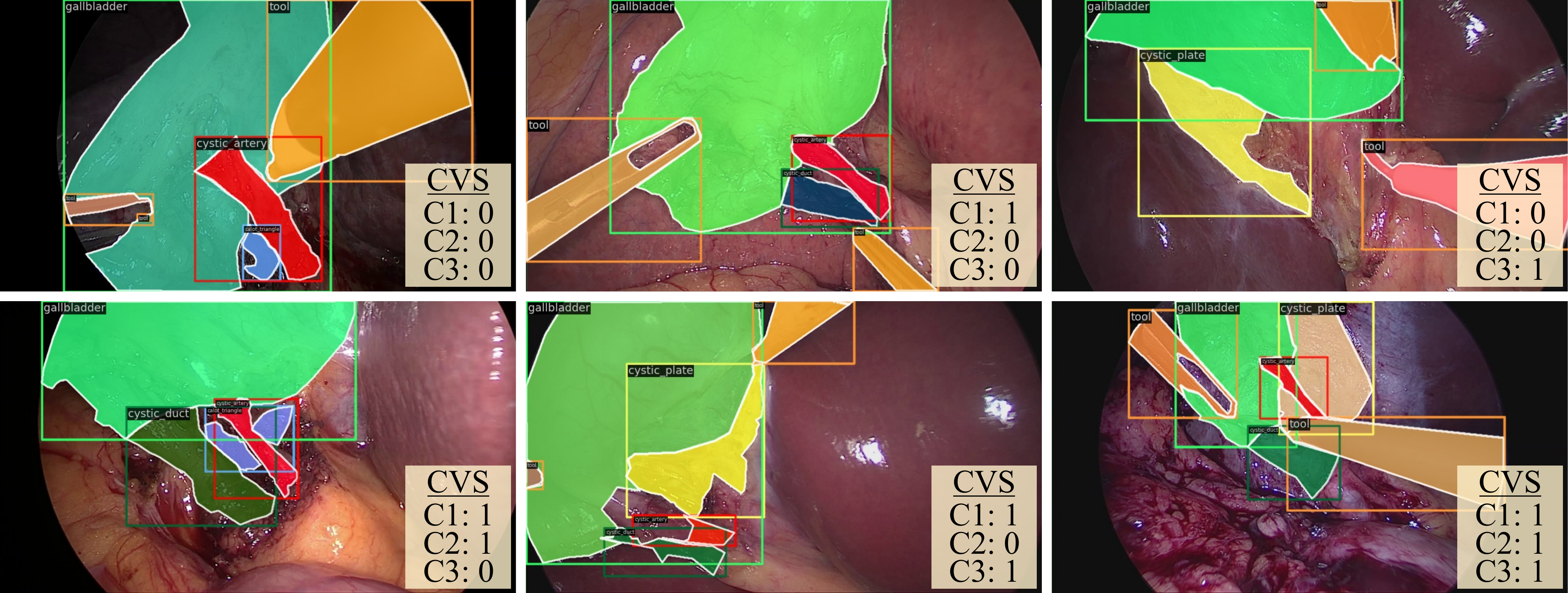}
\captionof{figure}{A series of example images from Endoscapes2023, illustrating the various types of annotations and representing a range of CVS achievement states.}
\label{fig:examples}
\end{figure*}

\begin{figure*}[!h]
\centering
\includegraphics[width=0.9\textwidth]{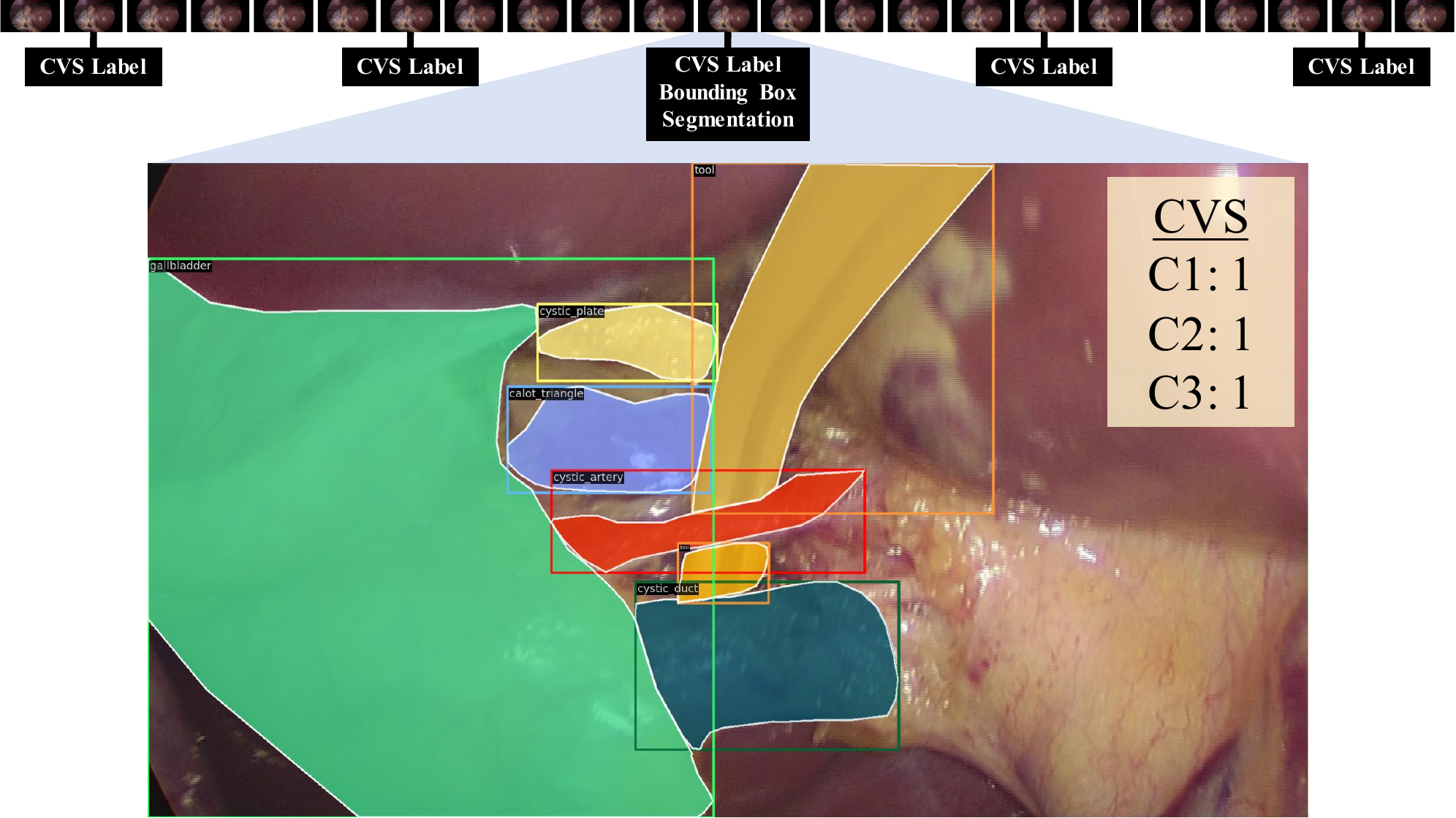}
\captionof{figure}{A visual summary of Endoscapes2023, representing the frequency and types of annotation in a sample video clip. The top row represents a sequence of frames, black boxes indicate available annotations, and frames without a black box are unlabeled.}
\label{fig:visual_summary}
\end{figure*}

\begin{multicols}{2}
\section{Introduction}

Automatic assessment of the critical view of safety (CVS) is an important problem in surgical data science that has been gaining interest in recent years~\cite{strasberg1995analysis,mascagni2020formalizing,mascagni2021artificial,murali2023latent,ban2023concept,rios2023cholec80,murali2023encoding}.
Unlike many surgical data science tasks that have been studied thus far (e.g. phase recognition, tool detection/segmentation), CVS assessment is particularly challenging because it relies on accurate and precise identification of fine-grained anatomical structures and concepts.
To enable research into this task, we present \textit{Endoscapes}, a dataset of laparoscopic cholecystectomy videos with annotations specifically targeted at tackling automated CVS assessment: namely, frame-level CVS annotations as well as segmentation/bounding boxes of critical anatomical structures/regions relevant for CVS assessment (such as the hepatocystic triangle, cystic artery/duct, and cystic plate).

In this report, we begin by describing the development of this dataset, which represents the culmination of several works~\cite{mascagni2021artificial,alapatt2021temporally,murali2023latent}.
Then, we provide a detailed enumeration of the contents of \textbf{Endoscapes2023}, released alongside this report, which comprises three sub-datasets categorized by annotation type and number of videos: Endoscapes-CVS201, Endoscapes-BBox201, and Endoscapes-Seg50.
Finally, we present benchmarks for three different tasks (object detection, instance segmentation, and CVS prediction).

\section{Context}

Automated CVS assessment was first tackled by our prior work DeepCVS~\cite{mascagni2021artificial}, where we present a method for joint anatomical segmentation and CVS prediction.
For this initial work, we hand-picked 2854 images where we deemed CVS assessable (in the anterior view) from 201 laparoscopic cholecystectomy (LC) videos, then annotated these images with each of the three CVS criteria (three binary image-level annotations), based on the protocol established in~\cite{mascagni2020formalizing}.
We also selected a subset of 402 images ($\sim$20\%), balancing optimal CVS achievement (3/3 criteria achieved) and suboptimal CVS achievement (0-2/3 criteria achieved) and further annotated them with segmentation masks.
This initial dataset, while instrumental in enabling a proof-of-concept study, was characterized by two critical limitations: (1) the frame selection process introduces bias during both training and evaluation, and (2) the dataset is very small, with a test set of only 571 images for CVS assessment and 80 images for segmentation.

Recognizing these issues, in a follow-up work~\cite{alapatt2021temporally}, we collected a vastly expanded dataset from the same 201 videos.
Here, rather than hand-selecting frames from the videos, we simply extract frames at 1fps from the region of video where CVS is adjudged to be evaluable\footnote{This region, which is a subset of the dissection phase, begins when ``any of the 3 CVS criteria is considered evaluable" and ends ``when the first clip is applied on the
cystic duct or the cystic artery", as defined in~\cite{mascagni2021surgical}.}, thereby eliminating selection bias.
For these frames, we then annotate segmentation (5 anatomical structures and 1 tool class) and CVS at regular intervals: 1 frame every 30 seconds for segmentation~\cite{alapatt2021temporally} and 1 frame every 5 seconds for CVS~\cite{murali2023latent}.
Removing the manual frame selection process introduced numerous issues during annotation, to which end we developed a more robust annotation protocol for both segmentation and CVS assessment~\cite{mascagni2021surgical}.
Finally, following this updated protocol and incorporating several rounds of review to ensure optimal consistency, we annotated 1933 total images with segmentation masks and 11090 images with CVS annotations from the 201 LC videos. The CVS annotations in particular were done independently by three expert clinicians to reduce bias; for evaluation, we define the ground-truth as the consensus of these annotations (majority vote for each criterion).
We additionally extract bounding box annotations automatically from the 1933 segmentation masks and augment the dataset with these labels~\cite{murali2023latent}.

\textbf{Endoscapes2023} is the result of all of these steps; in the following section, we comprehensively describe the dataset contents and characteristics. 

\section{Dataset}

\begin{table*}[t]
\captionof{table}{Detailed overview of the various subsets of Endoscapes2023.}
\resizebox{\textwidth}{!}{
\begin{tabular}{cccccccccccccc}
\multirow{2}{*}{\textbf{Subset}} & \multicolumn{4}{c}{\textbf{\# Videos}} & \multicolumn{4}{c}{\textbf{\# Frames}} & \multirow{2}{*}{\textbf{\begin{tabular}[c]{@{}c@{}}Annotation\\ Type\end{tabular}}} & \multicolumn{4}{c}{\textbf{\# Annotations}} \\
 & \textbf{Train} & \textbf{Val} & \textbf{Test} & \textbf{Total} & \textbf{Train} & \textbf{Val} & \textbf{Test} & \textbf{Total} &  & \textbf{Train} & \textbf{Val} & \textbf{Test} & \textbf{Total} \\ \hline
\textbf{Endoscapes-CVS201} & 120 & 41 & 40 & 201 & 36694 & 12372 & 9747 & 58813 & \begin{tabular}[c]{@{}c@{}}CVS (Binary\\ Multilabel)\end{tabular} & 6960 & 2331 & 1799 & 11090 \\
\textbf{Endoscapes-BBox201} & 120 & 41 & 40 & 201 & 36694 & 12372 & 9747 & 58813 & Bounding Box & 1212 & 409 & 312 & 1933 \\
\textbf{Endoscapes-Seg50} & 30 & 10 & 10 & 50 & 10380 & 2310 & 2250 & 14940 & \begin{tabular}[c]{@{}c@{}}Segmentation\\ (Instance +\\ Semantic)\end{tabular} & 343 & 76 & 74 & 493
\end{tabular}}
\label{table:all_datasets_and_splits}
\end{table*}

\begin{figure*}
\centering
\includegraphics[width=0.95\textwidth]{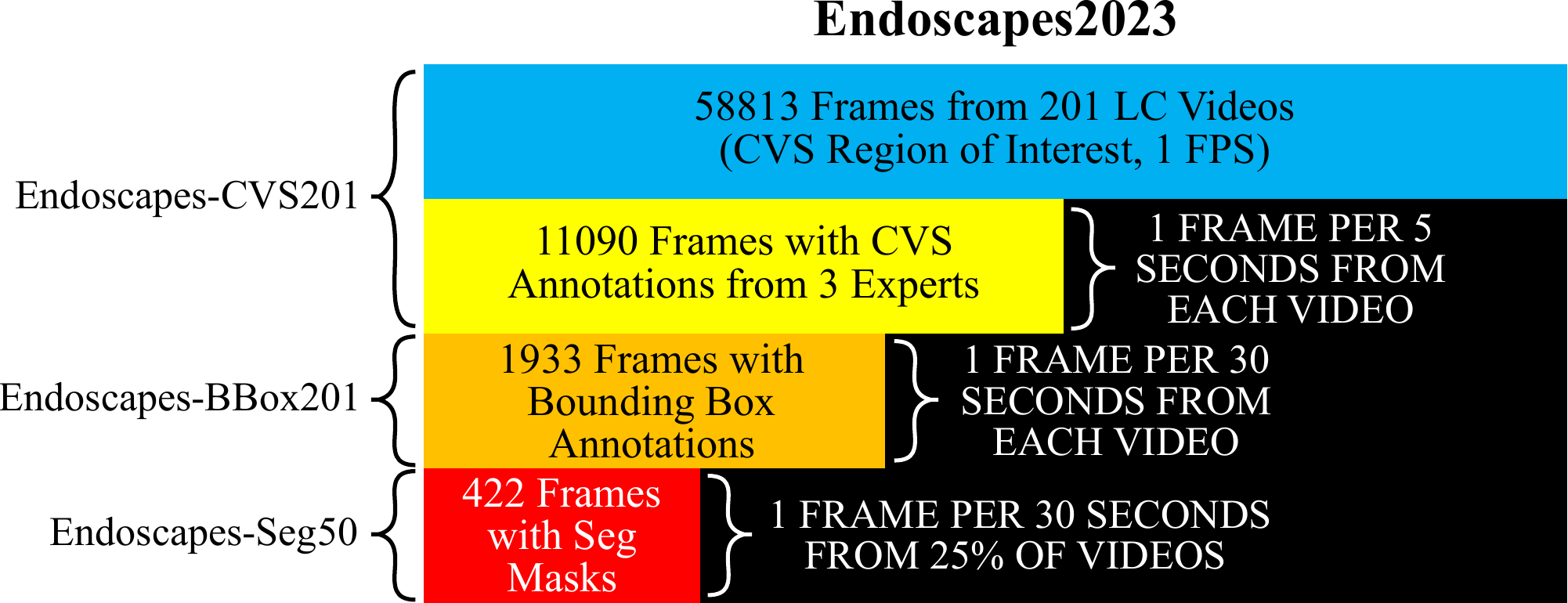}
\caption{A detailed enumeration of the contents of Endoscapes2023. Although Endoscapes-BBox201 and Endoscapes-Seg50 are only shown as containing paired frames and annotations, they also encompass all frames from the respective set of videos. Table \ref{table:all_datasets_and_splits} shows the total number of frames as well as number of annotated frames for each sub-dataset.}
\label{fig:details}
\end{figure*}

Figure \ref{fig:visual_summary} shows the data and annotations associated with a section of one LC video from Endoscapes2023, and illustrates our multifaceted annotation approach (semi-sparse for CVS, sparse for bounding boxes and segmentation masks).
Endoscapes2023 contains four subsets:

\noindent \textbf{Endoscapes-CVS201.} A collection of 58813 frames from the dissection phase (defined as in~\cite{twinanda2016endonet}) of the 201 LC videos.
Of these 58813 frames, 11090 frames (1 frame every 5 seconds) are annotated with CVS by three experts, where
CVS labels are a collection of three binary image-level annotations indicating achievement of each of the three CVS criteria: namely, C1 - Two Structures, C2 - Hepatocystic Triangle Dissection, and C3 - Cystic Plate, as formalized in~\cite{mascagni2020formalizing}.
Cohen’s kappa for inter-rater agreements for C1, C2, and C3, and the overall CVS assessment (1 if all 3 criteria are achieved, 0 otherwise) were $0.33$, $0.53$, $0.44$, and $0.38$.
These Cohen’s kappa values were first computed between each pair of annotators, then averaged across the three possible pairs for each of the three criteria and overall CVS.

We split the videos into 120 training, 41 validation, and 40 test, yielding 36694 training frames, 12372 validation frames, and 9747 testing frames; of these, 6960 training frames, 2331 validation frames, and 1799 testing frames contain CVS annotations.
To generate the splits, we employ stratified random sampling, using video-level CVS achievement for stratification; splitting the dataset by video (by patient) follows common practice in surgical video analysis~\cite{twinanda2016endonet,zisimopoulos2018deepphase,nwoye2022data}.
These same dataset splits were also used in~\cite{alapatt2021temporally,murali2023latent,murali2023encoding,alapatt2023jumpstarting}, thereby facilitating comparisons.

\noindent \textbf{Endoscapes-BBox201.} The same collection of 58813 frames as Endoscapes-CVS201 of which 1933 frames (1 frame every 30 seconds) are annotated with bounding boxes for 6 different classes (5 anatomical structures and a tool class).
Using the video splits from above yields 1212, 409, and 312 training, validation, and testing frames respectively.
Note that each fold (training, validation, testing) of Endoscapes-BBox201 is a strict subset of the respective fold in Endoscapes-CVS201.

\noindent \textbf{Endoscapes-Seg201.} The same as Endoscapes-BBox201 but with additional segmentation masks for the 1933 frames (1212 training, 409 validation, and 312 testing) that already contain bounding box annotations.
\textit{We only release a subset of Endoscapes-Seg201, called Endoscapes-Seg50 and described below, as part of Endoscapes2023.}

\noindent \textbf{Endoscapes-Seg50.} A subset of Endoscapes-Seg201 containing 14940 frames from 50 videos (a subset of the 201 videos), of which 493 (1 frame every 30 seconds) are annotated with segmentation masks.
We select the 50 videos by selecting 30, 10, and 10 videos from the aforementioned 120 training, 41 validation, and 40 test videos respectively, employing stratified random sampling based on video-level CVS achievement.
This ensures that Endoscapes-Seg50 can be used in conjunction with Endoscapes-BBox201 and Endoscapes-CVS201 to train mixed-supervision models for object detection, segmentation, and CVS prediction.
Following the splitting, we have 10380 training, 2310 validation, and 2250 testing frames of which 343, 76, and 74 respectively contain segmentation masks.
It should be noted that the 30 training videos of Endoscapes-Seg50 correspond to the third 25\% subset fold of Endoscapes-[CVS/Seg]201 that we use for low-label experiments in~\cite{alapatt2023jumpstarting,ramesh2023dissecting}.

\begin{table*}[t]
\caption{Performance of different object detectors on Endoscapes-BBox201.}
\label{table:object_detection}
\centering
\resizebox{0.8\textwidth}{!}{
\begin{tabular}{cccccccc}
\multirow{2}{*}{\textbf{Detector}} & \multicolumn{7}{c}{\textbf{Detection mAP@{[}0.5:0.95{]}}} \\
 & \textbf{\begin{tabular}[c]{@{}c@{}}Cystic\\ Plate\end{tabular}} & \textbf{\begin{tabular}[c]{@{}c@{}}HC Triangle\\ Dissection\end{tabular}} & \textbf{\begin{tabular}[c]{@{}c@{}}Cystic\\ Artery\end{tabular}} & \textbf{\begin{tabular}[c]{@{}c@{}}Cystic\\ Duct\end{tabular}} & \textbf{Gallbladder} & \textbf{Tool} & \textbf{Overall} \\ \hline
\textbf{Faster-RCNN} & 8.0 & 19.4 & 12.6 & 15.5 & 62.9 & 61.5 & 30.0 \\
\textbf{Cascade-RCNN} & 7.5 & 18.4 & 12.1 & 13.8 & 61.5 & 62.4 & 29.3 \\
\textbf{Deformable-DETR} & 9.3 & 19.9 & 14.9 & 19.0 & 69.6 & 63.7 & 32.7
\end{tabular}}
\end{table*}

\begin{table*}[t]
\caption{Per-class performance of different instance segmentation models trained and tested on Endoscapes-Seg50.}
\label{table:insseg_classwise}
\centering
\begin{tabular}{cccccccc}
\multirow{2}{*}{\textbf{Model}} & \multicolumn{7}{c}{\textbf{Detection mAP@{[}0.5:0.95{]} $|$ Segmentation mAP@{[}0.5:0.95{]}}} \\
 & \textbf{\begin{tabular}[c]{@{}c@{}}Cystic\\ Plate\end{tabular}} & \textbf{\begin{tabular}[c]{@{}c@{}}HC Triangle\\ Dissection\end{tabular}} & \textbf{\begin{tabular}[c]{@{}c@{}}Cystic\\ Artery\end{tabular}} & \textbf{\begin{tabular}[c]{@{}c@{}}Cystic\\ Duct\end{tabular}} & \textbf{Gallbladder} & \textbf{Tool} & \textbf{Overall} \\ \hline
\textbf{Mask-RCNN} & 2.8 $|$ 3.3 & 2.9 $|$ 3.8 & 12.7 $|$ 11.9 & 7.4 $|$ 7.9 & 45.8 $|$ 59.1 & 49.7 $|$ 51.2 & 20.2 $|$ 22.9 \\
\textbf{Cascade-Mask-RCNN} & 3.1 $|$ 6.5 & 11.1 $|$ 6.7 & 15.7 $|$ 10.8 & 11.7 $|$ 10.0 & 62.0 $|$ 62.7 & 63.2 $|$ 56.8 & 27.8 $|$ 25.6 \\
\textbf{Mask2Former} & 1.4 $|$ 1.7 & 14.3 $|$ 8.3 & 6.4 $|$ 7.6 & 14.7 $|$ 15.9 & 68.7 $|$ 62.6 & 67.1 $|$ 63.5 & 28.8 $|$ 26.6
\end{tabular}
\end{table*}

Our tiered annotation approach (a plethora of unlabeled images, many `cheap' CVS annotations, low-frequency intermediate cost bounding box annotations, and a small subset with expensive segmentation masks) not only effectively simulates realistic annotation budgets but also enables a huge diversity of experiments with regard to mixed-supervision, semi-supervision, and temporal modeling.

Figure \ref{fig:details} illustrates the different subsets of Endoscapes2023.
Meanwhile, Table \ref{table:all_datasets_and_splits} summarizes the details of each subset.

\subsection{Statistics}

Table \ref{table:cvs_dataset_stats} details the CVS achievement statistics -- per-criteria and overall -- at both frame- and video-level, calculated based on the expert consensus. Video-level achievement only requires achievement in at least one frame of the video; additionally, video-level CVS is considered achieved as long as each criterion is achieved at least once in the video, but not necessarily concurrently. Meanwhile, the frame-level overall CVS achievement requires achievement of all three criteria in the same frame. As a result, video-level achievement rates are much higher than the frame-level statistics.

\begin{center}
\captionof{table}{Achievement Rates (\%) of each CVS Criterion and Overall CVS in Endoscapes-CVS201. A criterion is achieved at the video-level if it is achieved in at least one frame of the video. Overall CVS is the maximum of the criteria-level achievements for each frame/video.}
\resizebox{0.48\textwidth}{!}{
\begin{tabular}{ccccccc}
\multirow{2}{*}{\textbf{Criterion}} & \multicolumn{3}{c}{\textbf{Frame}} & \multicolumn{3}{c}{\textbf{Video}} \\
 & \textbf{Train} & \textbf{Val} & \textbf{Test} & \textbf{Train} & \textbf{Val} & \textbf{Test} \\ \hline
\textbf{C1: Two Structures} & 15.6 & 16.3 & 24.0 & 55.8 & 53.7 & 62.5 \\
\textbf{\begin{tabular}[c]{@{}c@{}}C2: Hepatocystic\\ Triangle Dissection\end{tabular}} & 11.2 & 12.5 & 17.1 & 48.3 & 53.7 & 50.0 \\
\textbf{C3: Cystic Plate} & 17.9 & 16.7 & 27.1 & 50.0 & 48.8 & 57.5 \\ \hline
\textbf{Overall} & 5.3 & 7.3 & 7.7 & 28.3 & 36.6 & 35.0
\end{tabular}}
\label{table:cvs_dataset_stats}
\end{center}

\vspace{15pt}

Meanwhile, table \ref{table:det_dataset_stats} details the object detection/segmentation dataset statistics.
We report object occurrence frequency, which refers to the percentage of frames that contain at least one instance of a particular object class occurs; classes with low frequency present more difficulty to object detectors and segmentation models. 

\begin{center}
\captionof{table}{Proportion of frames (\%) containing each object class in Endoscapes-BBox201 and Endoscapes-Seg50. A frame is considered to contain the object class if it has at least one instance of that class.}
\label{table:det_dataset_stats}
\centering
\resizebox{0.48\textwidth}{!}{
\begin{tabular}{ccccccc}
\multirow{3}{*}{\textbf{Class}} & \multicolumn{6}{c}{\textbf{Dataset and Split}} \\
 & \multicolumn{3}{c}{\textbf{Endoscapes-BBox201}} & \multicolumn{3}{c}{\textbf{Endoscapes-Seg50}} \\
 & \textbf{Train} & \textbf{Val} & \textbf{Test} & \textbf{Train} & \textbf{Val} & \textbf{Test} \\ \hline
\textbf{Cystic Plate} & 35.7 & 33.7 & 41.7 & 42.3 & 38.2 & 24.3 \\
\textbf{\begin{tabular}[c]{@{}c@{}}HC Triangle\\ Dissection\end{tabular}} & 35.2 & 32.5 & 39.7 & 33.2 & 32.9 & 9.5 \\
\textbf{Cystic Artery} & 52.5 & 46.9 & 61.5 & 58.9 & 44.7 & 63.5 \\
\textbf{Cystic Duct} & 78.6 & 65.3 & 82.7 & 74.6 & 85.5 & 71.6 \\
\textbf{Gallbladder} & 96.9 & 88.0 & 92.0 & 97.7 & 96.1 & 78.4 \\
\textbf{Tool} & 94.2 & 86.1 & 88.8 & 93.9 & 96.1 & 74.3
\end{tabular}}
\end{center}

\section{Benchmark}
In this section, we present benchmarks for three tasks: (1) object detection using Endoscapes-BBox201, (2) instance segmentation using Endoscapes-Seg50 (or Endoscapes-Seg201), and (3) CVS prediction using Endoscapes-CVS201, Endoscapes-BBox201, and Endoscapes-Seg50 (or Endoscapes-Seg201).

\begin{center}
\captionof{table}{Performance of different instance segmentation models on Endoscapes-Seg50. We report the results when trained on the train set of Endoscapes-Seg50 as well as when trained on the private Endoscapes-Seg201 (ceiling). We report both bounding box and segmentation mAP, following common practice.}
\resizebox{0.48\textwidth}{!}{
\begin{tabular}{cccc}
\multirow{2}{*}{\textbf{Train Set}} & \multirow{2}{*}{\textbf{Model}} & \multicolumn{2}{c}{\textbf{mAP@{[}0.5:0.95{]}}} \\
 &  & \textbf{Bbox} & \textbf{Segm} \\ \hline
\multirow{3}{*}{\textbf{Endoscapes-Seg50}} & \textbf{Mask-RCNN} & 20.2 & 22.9 \\
 & \textbf{Cascade-Mask-RCNN} & 27.8 & 25.6 \\
 & \textbf{Mask2Former} & 28.8 & 26.6 \\ \hline
\multirow{3}{*}{\textbf{Endoscapes-Seg201}} & \textbf{Mask-RCNN} & 29.2 & 28.6 \\
 & \textbf{Cascade-Mask-RCNN} & 31.7 & 30.0 \\
 & \textbf{Mask2Former} & 30.9 & 31.2
\end{tabular}}
\label{table:instance_segmentation}
\end{center}

\subsection{Object Detection}

We present the performance of three different object detectors on the Endoscapes2023 test set in Table \ref{table:object_detection}: Faster-RCNN, Cascade-RCNN, and Deformable-DETR, originally reported in~\cite{murali2023latent}.
We initialize all detectors using COCO-pretrained weights.
Then, for Faster-RCNN and Cascade-RCNN, we train for 20 epochs, while for Deformable-DETR, we train for 30 epochs, following the COCO learning schedule for all detectors.
Following most literature in object detection, we report mAP@[0.5:0.95] and breakdown performance per object class.

\subsection{Segmentation}
For the two segmentation tasks, we report two performance numbers for each model: performance when trained and tested on Endoscapes-Seg50, as well as the performance of models trained on Endoscapes-Seg201 as in~\cite{alapatt2021temporally,murali2023latent,murali2023encoding}.
This latter performance number represents a ceiling performance that can inform and drive the development of label-efficient segmentation methods.
Table \ref{table:instance_segmentation} shows the performance of Mask-RCNN, Cascade-Mask-RCNN, and Mask2Former for instance segmentation, using segmentation mAP@[0.5:0.95] as the metric.
Then, we break down per-class performance in Table \ref{table:insseg_classwise}.

\begin{table*}[t]
\centering
\caption{CVS Prediction Performance Benchmark on the test set of Endoscapes-CVS201. Models are trained in three different settings, arranged from most label-efficient to least label-efficient.
Asterisks (*) indicate spatiotemporal methods. Performances for the first two settings (Endoscapes-CVS201 only and Endoscapes-BBox201 + Endoscapes-CVS201) are cited from~\cite{murali2023latent,murali2023encoding}.
Meanwhile, to represent the last experimental setting from~\cite{murali2023latent,murali2023encoding} that uses Endoscapes-Seg201, we rerun all models using Endoscapes-Seg50 to train the underlying Mask-RCNN model. The performance drop from the results cited in~\cite{murali2023latent} can be attributed to the use of far fewer segmentation masks during training.}
\label{table:cvs}
\resizebox{0.9\textwidth}{!}{
\begin{tabular}{ccccccccccc}
\multirow{2}{*}{\textbf{Train Labels}} & \multirow{2}{*}{\textbf{Model}} & \multicolumn{4}{c}{\textbf{CVS mAP}} &  & \multicolumn{4}{c}{\textbf{CVS Bacc}} \\
 &  & \textbf{C1} & \textbf{C2} & \textbf{C3} & \textbf{Avg} &  & \textbf{C1} & \textbf{C2} & \textbf{C3} & \textbf{Avg} \\ \hline
\multirow{3}{*}{\textbf{Endoscapes-CVS201 Only}} & \textbf{ResNet50} & 42.9 & 46.5 & 65.1 & 51.5 &  & 57.6 & 63.7 & 70.2 & 63.8 \\
 & \textbf{\begin{tabular}[c]{@{}c@{}}ResNet50 w/\\ Reconstruction\end{tabular}} & 46.2 & 48.1 & 62.8 & 52.1 &  & 67.3 & 64.3 & 60.7 & 64.1 \\
 & \textbf{ResNet50-MoCov2} & 46.4 & 56.5 & 69.4 & 57.4 &  & 63.5 & 68.1 & 68.4 & 66.7 \\ \hline
\multirow{7}{*}{\textbf{\begin{tabular}[c]{@{}c@{}}Endoscapes-BBox201 +\\ Endoscapes-CVS201\end{tabular}}} & \textbf{LayoutCVS} & 53.5 & 37.6 & 53.4 & 48.2 &  & 68.2 & 65.7 & 62.3 & 65.4 \\
 & \textbf{DeepCVS} & 53.7 & 44.9 & 52.9 & 50.5 &  & 68.7 & 69.3 & 63.4 & 67.1 \\
 & \textbf{ResNet50-DetInit} & 61.2 & 53.7 & 65.7 & 60.2 &  & 71.5 & 72.9 & 65.8 & 70.0 \\
 & \textbf{LG-CVS} & 65.7 & 56.3 & 67.9 & 63.3 &  & 75.8 & 76.1 & 72.9 & 74.9 \\
 & \textbf{DeepCVS-Temporal*} & 59.4 & 47.8 & 66.2 & 57.8 &  & 69.1 & 64.2 & 67.4 & 66.9 \\
 & \textbf{STRG*} & 53.8 & 57.3 & 70.4 & 60.5 &  & 64.8 & 69.2 & 75.4 & 69.8 \\
 & \textbf{SV2LSTG*} & 65.9 & 59.0 & 70.9 & 65.3 &  & 70.4 & 68.4 & 73.6 & 70.8 \\ \hline
\multirow{7}{*}{\textbf{\begin{tabular}[c]{@{}c@{}}Endoscapes-Seg50 +\\ Endoscapes-CVS201\end{tabular}}} & \textbf{LayoutCVS} & 62.2 & 50.5 & 58.5 & 57.1 &  & 68.2 & 70.8 & 69.2 & 69.4 \\
 & \textbf{DeepCVS} & 62.3 & 45.7 & 59.9 & 56.0 &  & 71.8 & 71.6 & 70.3 & 71.2 \\
 & \textbf{ResNet50-DetInit} & 58.5 & 54.9 & 62.0 & 58.5 &  & 72.2 & 70.0 & 69.2 & 70.5 \\
 & \textbf{LG-CVS} & 67.9 & 53.6 & 68.0 & 63.2 &  & 75.8 & 75.4 & 73.2 & 74.8 \\
 & \textbf{DeepCVS-Temporal*} & 63.5 & 53.8 & 64.4 & 60.5 & \multicolumn{1}{l}{} & 75.6 & 73.8 & 71.1 & 73.5 \\
 & \textbf{STRG*} & 52.4 & 55.5 & 69.2 & 59.0 & & 62.3 & 69.1 & 73.7 & 68.4 \\
 & \textbf{SV2LSTG*} & 66.0 & 56.4 & 70.6 & 64.3 & \multicolumn{1}{l}{} & 76.2 & 73.8 & 70.3 & 73.4 \\ \hline
 \multirow{7}{*}{\textbf{\begin{tabular}[c]{@{}c@{}}Endoscapes-Seg201 +\\ Endoscapes-CVS201\end{tabular}}} & \textbf{LayoutCVS} & 63.4 & 49.1 & 58.2 & 56.9 & & 69.4 & 67.6 & 68.5 & 68.5 \\
 & \textbf{DeepCVS} & 65.9 & 52.6 & 61.8 & 60.2 &  & 74.0 & 73.4 & 70.7 & 72.3 \\
 & \textbf{ResNet50-DetInit} & 57.3 & 54.9 & 69.3 & 60.5 & & 69.0 & 70.2 & 74.3 & 71.2 \\
 & \textbf{LG-CVS} & 69.5 & 60.7 & 71.8 & 67.3 & & 78.6 & 81.4 & 79.4 & 79.8 \\
 & \textbf{DeepCVS-Temporal*} & 67.8 & 55.9 & 67.7 & 63.8 & \multicolumn{1}{l}{} & 78.2 & 75.4 & 75.6 & 76.4 \\
 & \textbf{STRG*} & 54.7 & 57.9 & 72.5 & 61.7 & & 63.3 & 74.2 & 80.3 & 72.6 \\
 & \textbf{SV2LSTG*} & 69.3 & 64.8 & 73.6 & 69.7 & & 83.5 & 82.8 & 80.9 & 82.4 \\
\end{tabular}}
\end{table*}

\subsection{CVS Prediction}

Finally, we present baseline results for CVS prediction. \cite{murali2023latent} contains a detailed exploration of the various possible experimental settings for CVS prediction, along with a benchmark of several single-frame approaches for CVS prediction that can make use of bounding box/segmentation labels if available.
Meanwhile,~\cite{murali2023encoding} explores spatiotemporal methods for CVS prediction, again considering the same experimental settings, while~\cite{alapatt2023jumpstarting} explores label-efficient CVS prediction.
We summarize the three experimental settings below and present results for Endoscapes2023, quoting~\cite{murali2023latent},~\cite{murali2023encoding}, and~\cite{alapatt2023jumpstarting} where applicable and rerunning the methods when the dataset used diverges from Endoscapes2023.

The three experimental settings for CVS prediction are: (1) using Endoscapes-CVS201 alone, thereby using only relatively `cheap' CVS annotations, (2) using Endoscapes-CVS201 in addition to Endoscapes-BBox201, and (3) using Endoscapes-CVS201 in addition to Endoscapes-Seg50.
For the last setting, as for the segmentation benchmark, we rerun the methods from~\cite{murali2023latent} and~\cite{murali2023encoding} using only Endoscapes-Seg50 as opposed to Endoscapes-Seg201.
We additionally present the original results using Endoscapes-Seg201 + Endoscapes-CVS201 from~\cite{murali2023latent,murali2023encoding} to illustrate ceiling performance.

\noindent \textbf{Metrics.} Because CVS classification is an imbalanced multilabel classification problem, selecting the right performance metric is paramount.
Using traditional classification accuracy can be extremely misleading because naive classifiers that output 0 for all the classes can perform quite well due to the dataset's inherent class imbalance.
We propose to measure performance using mean average precision (mAP) and balanced accuracy (bAcc), the former capturing model discriminative capability and the latter summarizing overall classification effectiveness.
We compute these metrics by first computing the respective criterion-wise values across all frames in Endoscapes-CVS201 and then averaging the three values.

\subsection{Summary of Methods}

We provide a brief summary of all methods presented in this benchmark below, separating them based on the labels required for training.
We include both single-frame methods and spatiotemporal methods in our benchmark, presenting all results in Table \ref{table:cvs} and measuring mAP and balanced accuracy.
For clarity and ease of comparison, we label the spatiotemporal methods with an asterisk in both the method summaries and in Table \ref{table:cvs}.

\subsubsection{Endoscapes-CVS201 Only}
The following methods only make use of Endoscapes-CVS201 for training/evaluation; they are highly label-efficient, using only CVS labels.

\noindent \textbf{ResNet50.} We show the results of a vanilla ResNet50 classifier with the final fully-connected layer modified to output three scores corresponding to the CVS criteria (from~\cite{murali2023latent}).

\noindent \textbf{ResNet50 with Reconstruction.} The same as the above with an added auxiliary image reconstruction objective (from~\cite{murali2023latent}).

\noindent \textbf{ResNet50-MoCov2.} Same ResNet50 classifier but initialized using a MoCov2-pretrained ResNet50 backbone (corresponds to the best performing initialization from~\cite{alapatt2023jumpstarting}.

\subsubsection{Endoscapes-BBox201 + Endoscapes-CVS201}
\label{section:cvs_with_boxes}
The following methods make use of Endoscapes-BBox201 as well as Endoscapes-CVS201, but do not require segmentation masks.
They represent intermediate label-efficiency.

\noindent \textbf{LayoutCVS.} A Faster-RCNN object detector is trained using the subset of images with bounding box annotations. Then, for CVS prediction, the trained, frozen object detector is run on an image, the output boxes are reorganized into a $C$-dimensional, pixel-wise layout, and this layout is passed to a ResNet18 classifier to predict the three CVS criteria.
$C$ refers to the number of object classes in the dataset, which in this case is $6$.
Performance reported from~\cite{murali2023latent}.

\noindent \textbf{DeepCVS.} The same as LayoutCVS, but the layout is concatenated with the original image prior to classification. Performance reported from~\cite{murali2023latent}.

\noindent \textbf{ResNet50-DetInit.} A ResNet50 classifier with the backbone weights initialized from the trained object detector. Performance reported from~\cite{murali2023latent}.

\noindent \textbf{LG-CVS.} The method proposed in~\cite{murali2023latent}, which extracts per-object features based on the detected objects and builds a latent graph with object semantic and visual properties as well as inter-object relations.
This latent graph is then used to predict CVS. Performance reported from~\cite{murali2023latent}. 

\noindent \textbf{DeepCVS-Temporal*.} A spatiotemporal extension of DeepCVS proposed in~\cite{murali2023encoding} that computes ResNet18 embeddings of a sequence of $T$ frames and processes with a Transformer to predict CVS.
Performance when setting $T = 10$ is reported from~\cite{murali2023encoding}.

\noindent \textbf{STRG*.} A spatiotemporal graph-based approach originally proposed in~\cite{wang2018videos} for fine-grained action recognition and adapted for CVS prediction in~\cite{murali2023encoding}. Performance when using a clip size of $10$ is reported from~\cite{murali2023encoding}. 

\noindent \textbf{SV2LSTG*.} The method proposed in~\cite{murali2023encoding} entitled ``Encoding \textbf{S}urgical \textbf{V}ideos as \textbf{L}atent \textbf{S}patio\textbf{T}emporal \textbf{G}raphs", where the LG-CVS~\cite{murali2023latent} encoder is used to extract a series of single frame graphs that are then linked temporally and used to predict CVS with a graph neural network. Performance when using a clip size of $10$ is reported from~\cite{murali2023encoding}.

\subsubsection{Endoscapes-Seg50 + Endoscapes-CVS201}
We also report the results when running all the methods from Section \ref{section:cvs_with_boxes} using Endoscapes-Seg50 instead of Endoscapes-BBox201; for all methods, we train a Mask-RCNN instance segmentation model in place of the Faster-RCNN object detector.
This represents the most label-intensive experimental setting.
We also report the original results of~\cite{murali2023latent,murali2023encoding}, where each model is trained on Endoscapes-Seg201 + Endoscapes-CVS201, to represent ceiling performance.

\subsubsection{Low-Label Settings}

Lastly, in Table \ref{table:cvs_low_data} we show selected results on two lower training data budgets of Endoscapes-CVS201: 12.5\% and 25\%, reporting results from~\cite{alapatt2023jumpstarting}.
These experiments, and the 12.5\% data budget in particular, represent the \textbf{most} label-efficient evaluation setting, using a small subset of annotations and requiring neither bounding box nor segmentation annotations, both of which are more expensive to collect than CVS labels.
For each budget, we sample three different sets of videos (15/30 videos respectively) from the train set of Endoscapes-CVS201.
The results using the entire train set of Endoscapes-CVS201 are shown in the first section of Table \ref{table:cvs} (Endoscapes-CVS201 Only), and represent a ceiling performance for these experiments.
We release these splits along with the remainder of the dataset to facilitate comparisons with the methods presented here.

\begin{center}
\captionof{table}{CVS prediction results on the test set of Endoscapes-CVS201 when training on 12.5\% and 25\% of the training set of Endoscapes-CVS201 (no use of bounding boxes/segmentation masks). Standard deviation is reported across three different data samples of the original Endoscapes-CVS201 training set for each label setting. The third split of the 25\% subset includes the same 30 videos as Endoscapes-Seg50, albeit with higher annotation frequency.}
\label{table:cvs_low_data}
\centering
\begin{tabular}{cccc}
\textbf{\% Labels} & \textbf{Method} & \textbf{CVS mAP} & \textbf{CVS bAcc} \\ \hline
\multirow{2}{*}{\textbf{12.5}} & \textbf{ResNet50} & 33.8 $\pm$ 4.1 & 51.2 $\pm$ 2.7 \\
 & \textbf{ResNet50-MoCov2} & 38.9 $\pm$ 5.6 & 55.5 $\pm$ 4.3 \\
\multirow{2}{*}{\textbf{25}} & \textbf{ResNet50} & 40.5 $\pm$ 7.0 & 56.1 $\pm$ 6.2 \\
 & \textbf{ResNet50-MoCov2} & 48.8 $\pm$ 3.2 & 63.6 $\pm$ 3.3
\end{tabular}
\end{center}

\subsection{Training Details}
We follow the implementation and training settings of~\cite{murali2023latent,alapatt2023jumpstarting} using the mmdetection framework~\cite{chen2019mmdetection} for all reported methods.
CVS prediction models are trained with a weighted binary cross entropy loss, where the weight for each criterion is computed by inverse frequency balancing.
All object detection and instance segmentation models are initialized with COCO-pretrained weights and finetuned.

\section{Conclusion}
This technical report provides a detailed description of Endoscapes2023, a dataset for automated assessment of the critical view of safety containing three different levels of annotation.
The presented data collection and annotation process not only enables a myriad of potential experimentation but also simulates realistic constraints with respect to data acquisition and annotation costs.
Lastly, we define official dataset splits and present an extensive benchmark of recent methods across various experimental settings, with the hope of facilitating proper comparisons in future work.

\bibliographystyle{sn-basic}
\bibliography{main}

\begin{thebibliography}{16}
\providecommand{\natexlab}[1]{#1}
\providecommand{\url}[1]{{#1}}
\providecommand{\urlprefix}{URL }
\providecommand{\doi}[1]{\url{https://doi.org/#1}}
\providecommand{\eprint}[2][]{\url{#2}}

\bibitem[{Alapatt et~al(2021)Alapatt, Mascagni, Vardazaryan, Garcia, Okamoto, Mutter, Marescaux, Costamagna, Dallemagne, and Padoy}]{alapatt2021temporally}
Alapatt D, Mascagni P, Vardazaryan A, et~al (2021) Temporally constrained neural networks (tcnn): A framework for semi-supervised video semantic segmentation. arXiv preprint arXiv:211213815

\bibitem[{Alapatt et~al(2023)Alapatt, Murali, Srivastav, Mascagni, Consortium, and Padoy}]{alapatt2023jumpstarting}
Alapatt D, Murali A, Srivastav V, et~al (2023) Jumpstarting surgical computer vision. arXiv preprint arXiv:231205968

\bibitem[{Ban et~al(2023)Ban, Eckhoff, Ward, Hashimoto, Meireles, Rus, and Rosman}]{ban2023concept}
Ban Y, Eckhoff JA, Ward TM, et~al (2023) Concept graph neural networks for surgical video understanding. IEEE Transactions on Medical Imaging

\bibitem[{Chen et~al(2019)Chen, Wang, Pang, Cao, Xiong, Li, Sun, Feng, Liu, Xu et~al}]{chen2019mmdetection}
Chen K, Wang J, Pang J, et~al (2019) Mmdetection: Open mmlab detection toolbox and benchmark. arXiv preprint arXiv:190607155

\bibitem[{Mascagni et~al(2020)Mascagni, Fiorillo, Urade, Emre, Yu, Wakabayashi, Felli, Perretta, Swanstrom, Mutter et~al}]{mascagni2020formalizing}
Mascagni P, Fiorillo C, Urade T, et~al (2020) Formalizing video documentation of the critical view of safety in laparoscopic cholecystectomy: a step towards artificial intelligence assistance to improve surgical safety. Surgical endoscopy 34(6):2709--2714

\bibitem[{Mascagni et~al(2021{\natexlab{a}})Mascagni, Alapatt, Garcia, Okamoto, Vardazaryan, Costamagna, Dallemagne, and Padoy}]{mascagni2021surgical}
Mascagni P, Alapatt D, Garcia A, et~al (2021{\natexlab{a}}) Surgical data science for safe cholecystectomy: a protocol for segmentation of hepatocystic anatomy and assessment of the critical view of safety. arXiv preprint arXiv:210610916

\bibitem[{Mascagni et~al(2021{\natexlab{b}})Mascagni, Vardazaryan, Alapatt, Urade, Emre, Fiorillo, Pessaux, Mutter, Marescaux, Costamagna et~al}]{mascagni2021artificial}
Mascagni P, Vardazaryan A, Alapatt D, et~al (2021{\natexlab{b}}) Artificial intelligence for surgical safety: automatic assessment of the critical view of safety in laparoscopic cholecystectomy using deep learning. Annals of Surgery

\bibitem[{Murali et~al(2023{\natexlab{a}})Murali, Alapatt, Mascagni, Vardazaryan, Garcia, Okamoto, Mutter, and Padoy}]{murali2023encoding}
Murali A, Alapatt D, Mascagni P, et~al (2023{\natexlab{a}}) Encoding surgical videos as latent spatiotemporal graphs for object and anatomy-driven reasoning. In: International Conference on Medical Image Computing and Computer-Assisted Intervention, Springer, pp 647--657

\bibitem[{Murali et~al(2023{\natexlab{b}})Murali, Alapatt, Mascagni, Vardazaryan, Garcia, Okamoto, Mutter, and Padoy}]{murali2023latent}
Murali A, Alapatt D, Mascagni P, et~al (2023{\natexlab{b}}) Latent graph representations for critical view of safety assessment. IEEE Transactions on Medical Imaging pp 1--1. \doi{10.1109/TMI.2023.3333034}

\bibitem[{Nwoye and Padoy(2022)}]{nwoye2022data}
Nwoye CI, Padoy N (2022) Data splits and metrics for method benchmarking on surgical action triplet datasets. arXiv preprint arXiv:220405235

\bibitem[{Ramesh et~al(2023)Ramesh, Srivastav, Alapatt, Yu, Murali, Sestini, Nwoye, Hamoud, Sharma, Fleurentin et~al}]{ramesh2023dissecting}
Ramesh S, Srivastav V, Alapatt D, et~al (2023) Dissecting self-supervised learning methods for surgical computer vision. Medical Image Analysis p 102844

\bibitem[{R{\'\i}os et~al(2023)R{\'\i}os, Molina-Rodriguez, Londo{\~n}o, Guill{\'e}n, Sierra, Zapata, and Giraldo}]{rios2023cholec80}
R{\'\i}os MS, Molina-Rodriguez MA, Londo{\~n}o D, et~al (2023) Cholec80-cvs: An open dataset with an evaluation of strasberg’s critical view of safety for ai. Scientific Data 10(1):194

\bibitem[{Strasberg et~al(1995)Strasberg, Hertl, and Soper}]{strasberg1995analysis}
Strasberg SM, Hertl M, Soper NJ (1995) An analysis of the problem of biliary injury during laparoscopic cholecystectomy. Journal of the American College of Surgeons 180(1):101--125

\bibitem[{Twinanda et~al(2016)Twinanda, Shehata, Mutter, Marescaux, De~Mathelin, and Padoy}]{twinanda2016endonet}
Twinanda AP, Shehata S, Mutter D, et~al (2016) Endonet: a deep architecture for recognition tasks on laparoscopic videos. IEEE transactions on medical imaging 36(1):86--97

\bibitem[{Wang and Gupta(2018)}]{wang2018videos}
Wang X, Gupta A (2018) Videos as space-time region graphs. In: ECCV, pp 399--417

\bibitem[{Zisimopoulos et~al(2018)Zisimopoulos, Flouty, Luengo, Giataganas, Nehme, Chow, and Stoyanov}]{zisimopoulos2018deepphase}
Zisimopoulos O, Flouty E, Luengo I, et~al (2018) Deepphase: surgical phase recognition in cataracts videos. In: MICCAI, Springer, pp 265--272

\end{thebibliography}

\end{multicols}

\end{document}